\documentclass[10pt, a4paper]{article}
\usepackage{lrec}
\usepackage{multibib}
\newcites{languageresource}{Language Resources}
\usepackage{graphicx}
\usepackage{tabularx}
\usepackage{soul}
\usepackage{authblk}
\usepackage{times}
\usepackage{latexsym}
\usepackage[formats]{listings}
\newcommand{\ignore}[1]{}
\usepackage{graphicx}
\usepackage{wrapfig}
\usepackage{comment}
\usepackage{tipa}
\usepackage{amsmath,amssymb}
\usepackage{amsmath}
\usepackage{algorithmicx}
\usepackage{makecell}
\usepackage{algpseudocode}
\usepackage{cases}
\usepackage{wrapfig}
\usepackage{listings}
\usepackage{setspace}
\usepackage{tabularx}
\usepackage{enumitem}
\usepackage{url}
\usepackage{csquotes}
\usepackage{wrapfig}
\usepackage{pgfgantt}
\usepackage{color,hyperref}
\usepackage{caption}
\usepackage{verbatim}
\usepackage{subcaption}
 \usepackage{color,hyperref}
 \usepackage{times}
 \usepackage{xcolor}
 \usepackage{soul}
 \usepackage[utf8]{inputenc}
\usepackage{epstopdf}
\usepackage[utf8]{inputenc}

\usepackage{hyperref}
\usepackage{xstring}

\usepackage{color}

\title{From Spatial Relations to Spatial Configurations}

\name{Soham Dan \textsuperscript{1}, Parisa Kordjamshidi\textsuperscript{3,4}, Julia Bonn\textsuperscript{2}, Archna Bhatia\textsuperscript{3}, Jon Cai\textsuperscript{2}, 
\\ {\bf \large Martha Palmer\textsuperscript{2}}, {\bf \large Dan Roth\textsuperscript{1}}
}
\address{ \textsuperscript{1}University of Pennsylvania,  \textsuperscript{2}University of Colorado, Bolder,  \textsuperscript{3}IHMC, \textsuperscript{4}Michigan State University \\
         \{sohamdan,danroth\}@seas.upenn.edu
         , \{julia.bonn,jon.z.cai,martha.palmer\}@colorado.edu
          ,\\ kordjams@msu.edu, abhatia@ihmc.us\\}

\abstract{
Spatial Reasoning from language is essential \ignore{component} for natural language understanding. Supporting it requires a representation scheme that can capture spatial phenomena encountered in language as well as in images and videos. Existing spatial representations\ignore{schemata primarily focus on spatial relations with a limited number of arguments, but these} are not sufficient for describing spatial configurations used in complex tasks.\ignore{like navigation or collaborative object construction} This paper extends the capabilities of existing spatial representation languages and increases coverage of the semantic aspects that are needed to ground spatial meaning {of natural language text} in the world. \ignore{Moreover, o}Our spatial relation language is able to represent a large, comprehensive set of spatial concepts crucial for reasoning and is designed to support composition of static and dynamic spatial configurations.\ignore{ Our proposed formulation addresses both static and dynamic spatial configurations.and is able to represent most spatial concepts}\ignore{complex spatial entities, spatial object properties, frame-of-reference, path components, and motion.} We integrate \ignore{enable automatic processing of spatial language in terms of the proposed formulation by incorporating} this\ignore{our spatial relation} language with the Abstract Meaning Representation (AMR) annotation schema and present a corpus annotated by this extended AMR\ignore{We extend the Abstract Meaning Representation (AMR) annotation schema by incorporating our spatial relation language; thus enabling automatic processing of spatial language in terms of the proposed formulation \ignore{spatial relation language}}. To exhibit the applicability of our representation scheme,  we annotate text taken from diverse datasets  and  show  how  we  extend  the  capabilities  of  existing  spatial  representation  languages with fine-grained decomposition of semantics and  blend it seamlessly with AMRs of sentences and discourse representations as a whole. \ignore{To exhibit the applicability of our representation scheme, we annotate text taken from diverse datasets and show how we extend the capabilities of existing spatial representation languages to better accommodate both static and dynamic spatial relations, and to blend seamlessly with AMRs of sentences and discourse representations as a whole.}\\ \newline \Keywords{Semantics, Knowledge Discovery/Representation} }

\begin{document}

\maketitleabstract

\section{Introduction}
\begin{figure}
  \includegraphics[width=\linewidth]{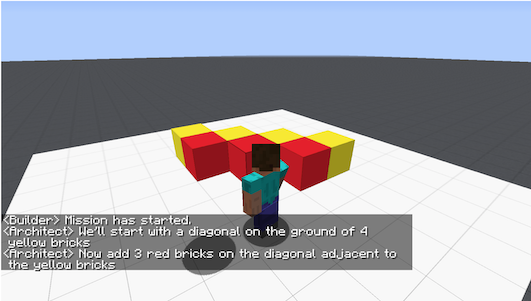}
  \caption{An instance of the collaborative building task. The last instruction was : \textit{Now add 3 red bricks on the diagonal adjacent to the yellow bricks.} }
  \label{fig:minecraft}
\end{figure}

Spatial reasoning is necessary for many tasks.
For example, consider the collaborative building task ~\cite{narayan-chen-etal-2019-collaborative} where a human architect (knows the target structure but cannot manipulate blocks) issues commands to a builder (who does not know the target structure but can place and manipulate blocks) with the aim of creating a complex target structure in a grounded environment (Minecraft). This task is challenging from a spatial reasoning perspective because the architect does not issue commands in terms of the unit blocks that the builder uses, but complex shapes and their spatial aspects. A single \ignore{natural language sentence}instruction can convey information about complex spatial configurations of several objects and describe nested relationships while shifting focus from describing one object to another. For example, in \textit{Place the red block on the yellow block which is to the left of the blue column.}, the focus shifts from describing the intended location of the red block to describing the location of the yellow block. Or, in a navigation description, there can be a sequence of spatial expressions describing a single object, as in \emph{Over the hill and through the woods...}. 
There is a severe limitation in the ability of existing spatial representation schemes to represent sentences involving complex spatial concepts. An example sentence from the Minecraft corpus is shown in Figure \ref{fig:minecraft}. 

Table \ref{tab:examples} highlights some of the challenging instances from a few recent datasets that cannot be adequately represented by existing spatial representations( for a more detailed comparison with previous schemes refer to Section 6). For instance, \ignore{~\cite{iso-space2011} does not represent the object properties in the first and third example and} the spatial representation in~\cite{LREC} does not capture the fine-grained  semantics of the object properties and does not distinguish between the frame of reference and the perspective. The scheme of ~\cite{iso-space2011} does not handle complex landmarks in the first example with the sequence of spatial properties.  We observe the ubiquity of these hard spatial descriptions across different datasets. Here is another such instance from NLVR ~\cite{suhr2017corpus} depicting \textit{spatial focus shift}: \textit{There is a box with a black item between 2 items of the same color and no item on top of that.} We propose an integrated view called \textsc{spatial configuration} that captures extensive spatial semantics that is necessary for reasoning. To develop an automatic builder, in the case of the collaborative construction task, we need an explicit representation of all involved  spatial components and their spatial relations in the instruction.

One of our key contributions is to represent this information compactly as a part of the configuration. Further, we can represent \textit{spatial focus shift} and \textit{nested concepts} as part of the scheme.  \ignore{We also situate the configuration scheme within the AMR framework as part of a spatial extension to AMR. Our spatial configuration scheme gives a general framework to represent spatial roles and relations and the extension of AMR with these roles gives a more expressive meaning representation language. }We also situate the configuration scheme within a new extended spatial AMR framework ~\cite{bonn2019spatial}. Our spatial configuration scheme takes the spatial roles and relations identified in spatial AMR graphs and converts them into an easy to read general framework, resulting in a maximally expressive spatial meaning representation language.

 Recent research shows the limitations of training monolithic deep learning models in two important aspects of  interpretability and generalizability~\cite{Hu2017LearningTR}.\ignore{Explicit and} Symbolic representations that can support reasoning capabilities\ignore{for learning models} are shown to have a critical role in improving both of the  above-mentioned aspects ~\cite{P18-1168,JamesSpQL,Suhr2017ACO}. A robust intermediate representation can help an off-the-shelf planner to ground a symbolic \ignore{meaning} representation of the spatial concepts to a final executable output. This representation should be as general and domain independent as possible to eliminate the need to retrain an automatic annotation system for a new domain.\ignore{that converts natural language input into the symbolic representation.} 
Using AMR ~\cite{banarescu2013abstract} as a stepping stone to our spatial configurations helps to ensure generalizability and domain independence.
Our goal is to identify various spatial semantic aspects that are required by a concrete meaning representation for downstream tasks. 

\begin{figure}
  \includegraphics[width=\linewidth]{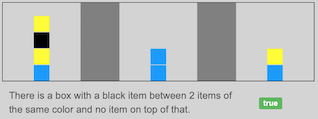}
  \caption{An example from the NLVR corpus that demonstrates \textit{spatial focus shift} from the \textit{black item} to the \textit{yellow item}. }
  \label{fig:nlvr}
\end{figure}

\ignore{\section{The need for a richer representation }
There is a severe limitation in the ability of existing spatial representation schemes to represent complex spatial structures as found in the collaborative building task ~\cite{narayan-chen-etal-2019-collaborative}. In this task, an architect(knows the target structure but cannot manipulate blocks) issues commands to the builder (does not know the target structure but can manipulate blocks) with the aim of creating the target structure in the Minecraft environment). An example sentence from this corpus is shown in Figure \ref{fig:minecraft}. Existing spatial representation schemes struggle in such cases where there are multiple landmarks but still one target location of placement. One of our contributions is to represent these information compactly as a part of the configuration. Further, we can represent as part of the scheme \textit{spatial focus shift} and \textit{nested concepts}. We observe the ubiquity of these hard spatial descriptions across different datasets. Here is another such instance from NLVR ~\cite{suhr2017corpus} : There is a box with a black item between 2 items of the same color and no item on top of that. Our spatial configuration scheme gives a general framework to represent spatial roles and relations and the extension of AMR with these roles gives a strictly more expressive meaning representation language. 
}
\section{Expanded AMRs} 
In this section, we briefly describe the spatial AMR extension that we designed to produce graphs suitable for mapping directly onto our spatial configurations. AMR is a good starting point for this annotation because it is domain general and relatively syntax-independent. The graphs themselves are comprised of nesting relations and concepts that may be inverted or rearranged depending on focus, which is useful in representing the kinds of complex and nesting spatial relations described above. While the new AMRs cover a richer set of spatial semantics and complex spatial relationships, they also still represent any non-spatial portions of the sentence. One contribution of our Spatial Configuration schema is the ability to extract the spatial relationships and spatial object properties from the AMR and organize them into an easily interpretable format that maintains the complex relationships and nesting.

\noindent The corpus that we describe here \footnote{\href{https://github.com/jbonn/CwC-Minecraft-AMR}{https://github.com/jbonn/CwC-Minecraft-AMR}} %\jb{what's the correct way to word citation of our other paper?}
 includes  annotation on over 5000 dialogue sentences (170 full dialogues) that discuss collaborative construction events in the Minecraft dataset ~\cite{narayan-chen-etal-2019-collaborative}. We have an additional 7600 annotated automatically-generated sentences representing builder actions, giving a total of 12,600 spatial AMRs. The AMR expansion involves the addition of 150+ new or updated PropBank~\cite{palmer2005proposition,gildea2002necessity} rolesets as well as a dozen general semantic frames, roles, and concepts that signal spatial relationships not previously covered by AMR. Using rolesets to annotate spatial relations allows us to disambiguate senses and to group together etymologically-related relations from different parts of speech within those senses. For example, the roleset {\bf align-02} includes aliases align-v, aligned-j, line-up-v, and in-line-p, and {\bf down-03} includes aliases down-p, down-r, downward-r, etc.. Prepositional and adverbial aliases are new for PropBank and AMR. The roles that make up the rolesets cover both semantically and pragmatically derived information and are labeled with a new set of spatial function tags that map to the elements of our Spatial Configurations. A typical directional spatial roleset includes two primary spatial entities held in comparison to one another (SE1, SE2), an axis or directional line between them (AXS), and an anchor which is used to name the entity whose spatial framework is being activated by the relation (ANC):

\noindent {\bf above-01:} higher on a vertical axis

:ARG1-SE1 entity above

:ARG2-SE2 entity below

:ARG3-ANC anchor

:ARG4-AXS axis

\noindent SE1 and SE2 are used for primary spatial entities that have an external~\cite{thora2011} relationship to each other; when the entities have an internal relationship, we create a separate sense and call them PRT and WHL. For example, {\bf top-06} entails an internal relationship (the trajector is part of the landmark) and {\bf on-top-03} entails an external relationship (the trajector and landmark are discrete from one another). Other relevant function tags are for angles (ANG), orientations (ORT), and scale (SCL) for use with scalar adjectives. 

\noindent New general semantic frames and roles allow us to target Spatial Configurations that are triggered outside of standard roleset applications. {\bf Have-configuration-91/:configuration} replaces :mod/:domain roles in cases where a set of entities is arranged into some sort of constellation, for example, \textit{the chairs are in a ring around the fire pit}. {\bf Spatial-sequence-91} provides a means of indicating that a temporal sequence of entities is meant to be interpreted as a spatial sequence, as in (\textit{put down red red green space red}) and \textit{now a green row of 3. Next a green row of 4.} {\bf Cartesian-framework-91} includes arguments for up to 3 axes, an origin, and a spatial entity housing the framework. {\bf Cartesian-coordinate-entity} is an entity frame whose arguments are :x, :y, :z and :framework. Other general semantic roles (with expected reified frames) include {\bf :orientation}, {\bf :size}, {\bf :color}, {\bf :anchor}, {\bf :axis}, and {\bf :pl}. {\bf :Pl} takes '+' as a value to indicate plurality, which AMR omitted and which is vital in grounded spatial annotation. 

\noindent The recently added prepositional and adjectival spatial relations are prevalent in language. For example, in the NLVR ~\cite{suhr2017corpus} training dataset (1000 sentences) consisting of 10,710 tokens, we found 1563 occurrences of these newly added frame file concepts, which is almost 15\% of all tokens, and in the 3\textsc{d blocks}\ignore{3D BLOCKS} dataset~\cite{bisk2018learning}, we found 39,917 occurrences of the new additions among a total of 250,000 tokens, almost 16\% of all tokens. The recent GQA dataset~\cite{hudson2019gqa} has 22.4\% spatial questions, 28\% object attribute questions and 52\% questions on compositionality. Our spatial representation scheme with modular configurations that facilitate compositionality for reasoning and explicit representation of object properties will be helpful for such tasks too.
%\ab{BTW, this is just about 6.85\% tokens, may be we can mention this in terms of sentences involving spatial concepts represented by these spatial prepositions and adjectives. I'm sure that will be a higher percentage, also anyway we are not just focused on the lexical items but spatial configurations which are instantiated by/in sentences.}

\begin{table*}[h]
   \small
   
    \centering
    \begin{tabular}{|p{1.5cm}|p{15cm}|}
    \hline
    Dataset & Example Sentence \\
    \hline
    BLOCKS & Texaco should line up on the top right corner of BMW in the middle of Adidas and HP (1) \\
    GQA & Are there any cups to the left of the tray that is on top of the table? (2)\\
    NLVR & There is a black square touching the wall with a blue square right on top of it. (3) \\
    SpaceEval & While driving along the village’s main road the GPS showed us the direction right ahead, and after two minutes, just few hundred meters from the end of the village, we reached the spot. (4)
    %there is a white cathedral with two narrow towers behind it and mountains in the background.
    \\
    SpRL & a white bungalow with big windows, stairs to the left and the right, a neat lawn and flowers in front of the house (5) \ignore{and trees at the back}  \\
    Minecraft & place a green block on the edge three spaces from the left side (6) \\
    \hline
    \end{tabular}
    \caption{Examples from popular datasets that highlight the need for a more expressive spatial representation language}
    \label{tab:examples}
\end{table*}

\begin{table*}[]
   \small
   
    \centering
    \begin{tabular}{|l|l|}
    \hline
     Trajector \textit{(tr)} & The entity whose location or trans-location is described in a spatial configuration. \\
     Landmark \textit{(lm)} &
     The reference object that describes the location of the \textit{tr} or is a part of its path of motion.
     %The reference object that describes the location of the \textit{tr} or the path of a motion.\ab{maybe we don't mention path since path is separately also mentioned below?!} 
     %\JB{maybe we could say "The reference phrase that spatially describes the tr"? Calling it an object is very confusing when it can also refer to complex phrases like "the street from home to the caffe", and 'location' is conceptually limiting}
     \\
     Motion Indicator\textit{(m)}& Spatial movement usually described by a motion verb. %\ab{can you check if you agree with the change here, motion is not really a verb but a movement that is described by a verb.}
     \\ %It is usually a motion verb that describes spatial movement.\\
     Frame-of-Ref.\textit{(FoR)}& A coordinate system to identify location of an object, can be intrinsic, relative or absolute.%\ab{added a description} 
     %JB\{how about: An anchored coordinate system that grounds the relation between tr and lm: intrinsic, relative, absolute.}
     \\
     Path \textit{(path)}& The $tr$'s location can be described via a path of motion instead of a basic $lm$. \\
     Viewer \textit{(v)}& When FoR is relative, this indicates the viewer as first, second or third person.\\
     Spatial Indicator \textit{(sp)}& The lexical form of the relation between the trajector and landmark. \\
     Qualitative type \textit{(QT)}& The qualitative/formal type of the relation between the $tr$ and $lm$. \\
     \hline
    \end{tabular}
    \caption{The components of a generic spatial configuration}
    \label{tab:spatialElements}
\end{table*}

\begin{figure*}[htp]
\centering
\begin{minipage}[b]{\textwidth}
\includegraphics[width=\textwidth]{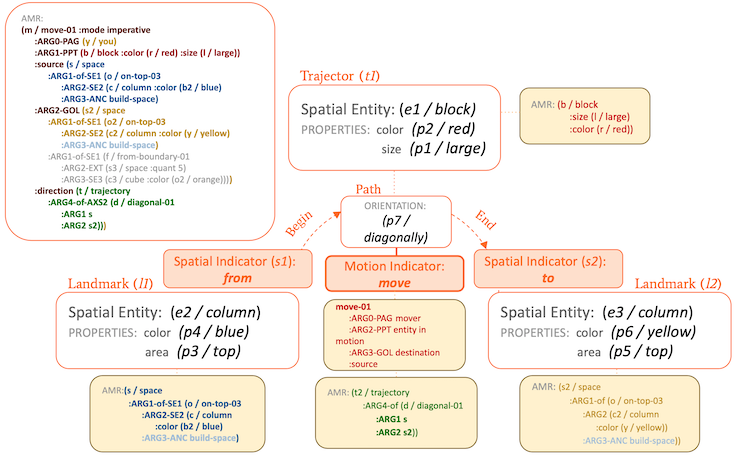}
\caption{Graphical Representation of Configuration 1 of Table \ref{eg-tower-tab} with aligned AMR : \textit{\textbf{Move the large red block diagonally from the top of the blue column to the top of the yellow column}, which is 5 spaces from the orange cube.}}\label{eg-tower-amr1}
\end{minipage}\qquad
\begin{minipage}[b]{\textwidth}
\includegraphics[width=\textwidth]{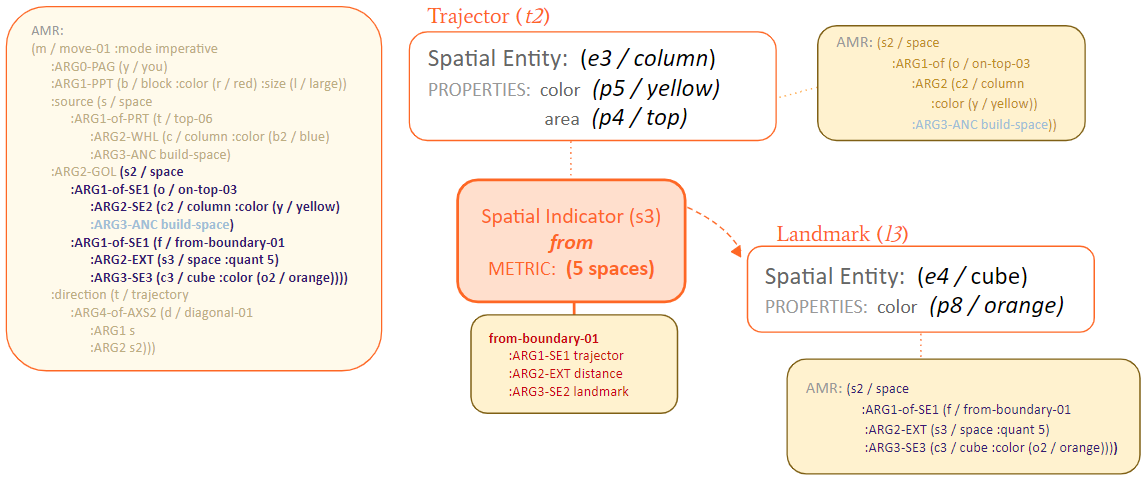}
\caption{Graphical Representation of Configuration 2 of Table \ref{eg-tower-tab} with aligned AMR : \textit{Move the large red block diagonally from the top of the blue column to the top of the yellow column, \textbf{which is 5 spaces from the orange cube.}}}\label{eg-tower-amr2}
\end{minipage}

\end{figure*}

\begin{figure*}[htp]
\centering
\begin{minipage}[b]{\textwidth}
\includegraphics[width=\textwidth]{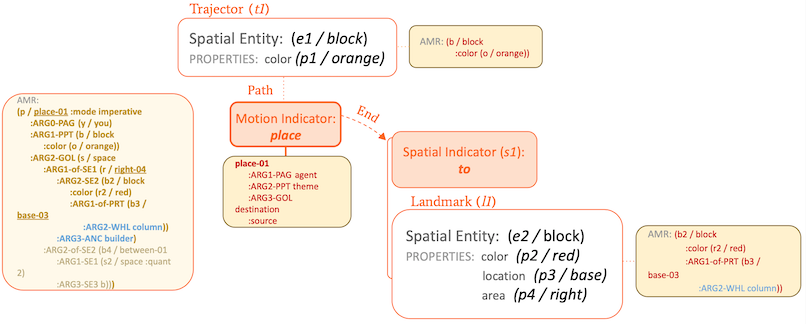}
\caption{Graphical Representation of Configuration 1 of Table \ref{eg-heart-tab} with aligned AMR  : \textit{\textbf{Place an orange block to the right of the base red block} with two spaces in between.}}\label{eg-heart-amr1}
\end{minipage}\qquad
\begin{minipage}[b]{\textwidth}
\includegraphics[width=\textwidth]{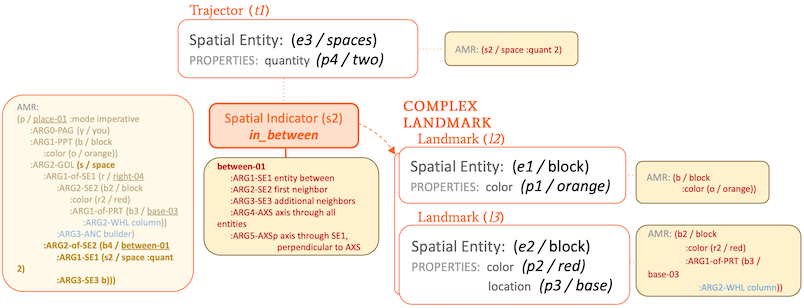}
\caption{Graphical Representation of Configuration 2 of Table \ref{eg-heart-tab} with aligned AMR : \textit{Place an orange block to the right of the base red block \textbf{with two spaces in between.}}}\label{eg-heart-amr2}
\end{minipage}

\end{figure*}

\newcolumntype{R}{>{\RaggedRight\arraybackslash}X}

\begin{table}[htbp]
    
    \centering
    \small
    %\begin{tabular}{ll}
    \begin{tabular}{|c|c|p{2.5cm}|}
                    
                     \hline
                    \multicolumn{3}{|c|}{Spatial Entities} \\ \hline
                    id & head & \makecell{properties \\\{ \textit{name = span} \}}  \\ \hline
                    e1 & block& \makecell{\{ size=large, \\  $ col = red  \} $} \\ \hline
                    e2& column & \makecell{\{ area=top,\\$ col = blue  \} $} \\ \hline
                    e3 & column & \makecell{\{ area=top,\\ $ col = yellow  \} $} \\ \hline
                    e4 & cube & \makecell{\{ col=orange  \}} \\ \hline  
                \end{tabular}

    \begin{tabular}{|c|p{3.2cm}|p{3.1cm}|} 
                    \hline
                     & \makecell{Configuration 1} & \makecell{Configuration 2}\\ \hline
                    tr&\makecell{$<t1, e1>$} & \makecell{$<t2, e3>$}\\ \hline
                    lm&$<l1,e2>,<l2,e3>$&\makecell{$<l3, e4>$}\\ \hline
                    %Lm&$<l2, e3>$\\ \hline
                    sp& \makecell{\textless \textit{s1,from} \textgreater \\ \textless \textit{s2,to} \textgreater} & \makecell{\textless \textit{s1,from,} \\  \{{metric=5 spaces} \} \textgreater} \\ \hline
                    %Sp&s2=to, []\\ \hline
                    m& \makecell{\textless \textit{m1,move,}  \textgreater} &\makecell{NULL}\\ \hline
                    path& \makecell{$<l1,s1,begin>$ \\ $<l2,s2,end>$ \\\{ \textit{orientation = diagonally} \}} &\makecell{NULL}  \\ \hline   
                    FoR & \makecell{$<l1,relative>$  \\ $<l2,relative>$ }& \makecell{$<l3,relative>$}\\ \hline
                    v& \makecell{first-person} & \makecell{first-person}\\ \hline  
                    QT& \textit{\textless directional, relative\textgreater}& \makecell{\textit{\textless distal, quantitative\textgreater}\\\textit{\textless topological, DC\textgreater}}
                    \\ \hline
                \end{tabular}
   % \end{tabular}
    \caption{Relational representation of spatial entities and configurations in the sentence: \emph{Move the large red block diagonally from the top of the blue column to the top of the yellow column, which is 5 spaces from the orange cube.}}
    \label{eg-tower-tab}
    
\end{table}

\newcolumntype{R}{>{\RaggedRight\arraybackslash}X}

\begin{table}[htbp]
    
    \centering
    \small
    %\begin{tabular}{ll}
    \begin{tabular}{|c|c|p{2.5cm}|}
                    
                     \hline
                    \multicolumn{3}{|c|}{Spatial Entities} \\ \hline
                    id & head & \makecell{properties \\ \{ \textit{name = span} \}}  \\ \hline
                    e1 & block& \makecell{\{ col=orange \}} \\ \hline
                    e2& block & \makecell{\{ loc=base,\\ $area=right$,\\$ col= red \} $} \\ \hline
                    e3 & spaces & \makecell{\{ quantity=2 \}} \\ \hline
                \end{tabular}
         \newline

        \begin{tabular}{|c|p{3.2cm}|p{3.1cm}|} 
                    \hline
                     & \makecell{Configuration 1} & \makecell{Configuration 2}\\ \hline
                    tr&\makecell{$<t1, e1>$} & \makecell{$<t2, e3>$}\\ \hline
                    lm&\makecell{$<l1,e2>$}&\makecell{$<l2, e2>,<l3, e1>$}\\ \hline
                    %Lm&$<l2, e3>$\\ \hline
                    sp& \makecell{ \textless \textit{s1,to} \textgreater } & \makecell{ \textless \textit{s2,in\_between} \textgreater } \\ \hline
                    %Sp&s2=to, []\\ \hline
                    m& \makecell{\textless \textit{m1,place}  \textgreater} &\makecell{NULL}\\ \hline
                    path& \makecell{$<implicit,begin>$ \\ $<l1,s1,end>$ } &\makecell{NULL}  \\ \hline   
                    FoR & \makecell{$<l1,relative>$}& \makecell{$<l2,relative>$ \\ $<l3,relative>$ }\\ \hline
                    v& \makecell{first-person} & \makecell{first-person}\\ \hline  
                    QT& \textit{\textless directional, relative\textgreater}& \makecell{\textit{\textless topological, EC\textgreater}}
                    \\ \hline
                \end{tabular}    
    
   % \end{tabular}
    \caption{Relational representation of spatial entities and configurations in the sentence: \emph{Place an orange block to the right of the base red block with two spaces in between.}}
    \label{eg-heart-tab}
    
\end{table}

\ignore{
\begin{table*}
       \begin{verbatim}
Move the large red block diagonally from the top of the blue column
to the top of the yellow column, which is 5 spaces from the orange cube.
SpRL
Move = motion indicator = M1
TR(the large red block) = T1
from the top of = spatial-indicator = I1
LM(the blue column)=L1, path = begin
to the top of= sparial-indicator = I2
LM= ( the yellow column) = L2, path = end
TR= ( the yellow column)= Tr2 
from = spatial-indicator = I3
LM= (Orange cube) = L3, path= begin
spatial-relation(T1, I1, L1, M1) S1
spatial-relation(T1, I2, L2,M1) S2
spatial-relation(Tr2, I3, L3, null) S3
S1(generaltype= topology, specific-type=RCC, value=PP, FoR=absolute)

\end{verbatim}
\end{table*}
\begin{table*}
       \begin{verbatim}
       Spatial Configuration
    sp-entity(the large red block)
    
\end{verbatim}
\end{table*}

\begin{table*}
       \begin{verbatim}
       ISO-space: the properties are represented but not anchored  
    sp-entity(the large red block, FAC)
    motion (move, ...)
\end{verbatim}
\end{table*}

\begin{table*}
       \begin{verbatim}
     AMR
    predicate(move, move-01, mode imperative) M1
    arg0(you,y) 
    arg1(block,a1_r)
    color(a1_r, red)
    size(a1_r, large)
    arg2(column,a2_b)
    color(a2,yellow)
    predicate(on top of, on-top-3,o2)
    (m1, a0, a1,a2)
    source(column, c2_b)
    color(c2_b, blue)
    path(c2_b,c_y, diagonal-01, "diagonally")
\end{verbatim}
\end{table*}

A circle closely touching a triangle. %\jb{what is this?}

What is ignored is the spatial properties annotation. Motion properties annotation. View annotation. The separate spatial entity is not annotated and multiple roles of TR and LM are allowed for a single entity. 
}

\section{Representation Language}
\begin{figure}[htp]
\centering
\begin{minipage}[b]{.45\textwidth}
\includegraphics[width=\textwidth]{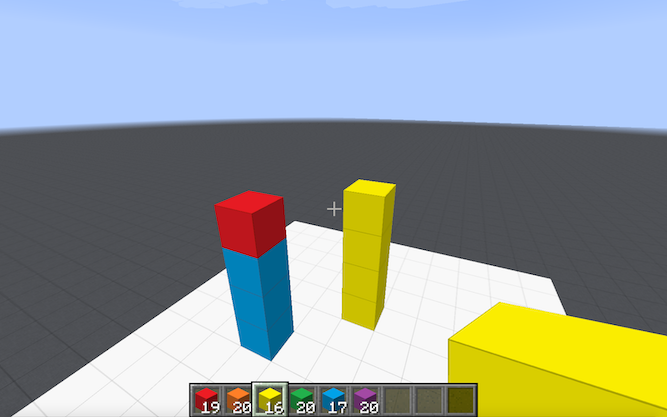}
%\caption{Before}\label{label-a}
\end{minipage}\qquad
\begin{minipage}[b]{.45\textwidth}
\includegraphics[width=\textwidth]{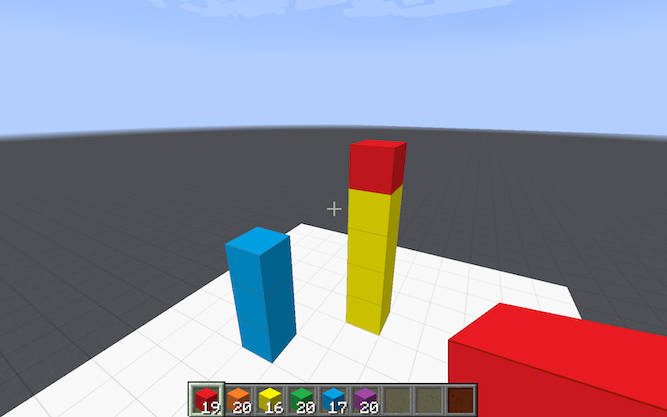}
%\caption{After}\label{label-b}
\end{minipage}
\caption{Move the large red block diagonally from the top of the blue column to the top of the yellow column ...}
\label{eg-tower-fig}
\end{figure}

\begin{figure}
	\centering
	\includegraphics[width=0.5\textwidth]{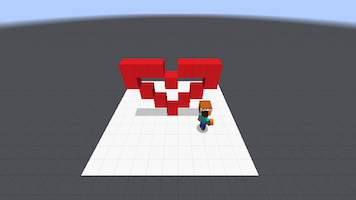}
	\caption
	{Place an orange block to the right of the base red block with two spaces in between.}
	\label{eg-heart-fig}
\end{figure}

In this section we describe the \textsc{spatial configuration} schema and as running examples we present two of our annotated instruction sentences from the Minecraft domain. We also include general examples beyond Minecraft to show our representation language is applicable universally. In Figures \ref{eg-tower-fig} and \ref{eg-heart-fig} we show the screenshots from the Minecraft environment of the actual execution of the instruction. In Tables \ref{eg-tower-tab} and \ref{eg-heart-tab} we show the detailed annotation of the spatial configuration elements. In Figures \ref{eg-tower-amr1}, \ref{eg-tower-amr2}, \ref{eg-heart-amr1} and \ref{eg-heart-amr2} we show the spatial configuration annotation graph and the aligned extended AMR annotations.

We represent each sentence $S$ as a set of \textit{spatial entities} $\{E\}$ and a set of \textit{configurations} $\{C\}$, that is, ${S=<\{E\},\{C\}>}$.  Each of our examples have two configurations because there is a shift of spatial focus; Tables \ref{eg-tower-tab} and \ref{eg-heart-tab} show them in detail.
Each {\bf spatial configuration} $C$ is described as a tuple, 

    \textit{C= \textless tr,\{lm \textpipe path\},m,sp,FoR,v,{QT} \textgreater}.
The elements of the tuple are described in Table \ref{tab:spatialElements}. 

\ignore{
Each {\bf spatial entity} $E$ is described as a tuple, $E=<id, head, \{prop\}>$
 where each $head$ corresponds to the entity (the actual shape in the Minecraft domain) that participates in spatial configurations with varied roles of $tr$, $lm$ or a part of a $path$. Spatial entities are described by {\bf spatial properties}, $prop$. We use these properties in the extended AMR that we develop. In Figures \ref{eg-tower-amr1}, \ref{eg-tower-amr2}, \ref{eg-heart-amr1} and \ref{eg-heart-amr2}  we show both the presented example with the corresponding AMR. We can see the tight connection between elements of the spatial configuration and the relevant portions of the AMR and how we fit the spatial concepts as a part of the larger AMR scheme to obtain a richer representation language.
}

\subsection{Spatial Entities and Properties} 
Each {\bf spatial entity} $E$ corresponds to an object in the text and can be described by a number of properties. We represent each spatial entity as follows: 
$E= <id, head, \{prop\}>$ where $id$ is an identifier and $head$ corresponds to the entity (the actual shape in the Minecraft domain) that participates in spatial configurations with varied roles of $tr$, $lm$ or a part of a $path$. Spatial entities are described by {\bf spatial properties}, $\{prop\}$. Each element in $prop$ is a  property. The properties are represented as 
%head refers to the actual sub-string in the sentence that is referring to the entity, and 
$prop=<name, span>$ where name takes a set of values, depending on the domain. In  Tables \ref{eg-tower-tab}, \ref{eg-heart-tab}  $\{shape, size, color, part-of, area, location, quantity\}$ are the properties of our domain and span refers to the actual lexical occurrence of the property. These properties are present as modifiers in the AMR graphs, which represent them via rolesets or general semantic roles like \textbf{:color} and \textbf{:size}. Notice that \textit{top} can be a \textit{part-of} property , as in, \textit{the top of the tower is touching the roof} or it could be a \textit{area} property , as in, \textit{the book is on top of the table}, where \textit{top} refers to the area on top of the table. \ignore{The location property describes the position of the trajector (for example, \textit{base} in Figure \ref{eg-heart-amr2}).} Spatial entity heads and their properties are all \textit{consuming tags}, that is, $head$ and $prop$ refer to the actual sub-strings in the sentence that are referring to the entity, or property, respectively.
In Figures \ref{eg-tower-amr1}, \ref{eg-tower-amr2}, \ref{eg-heart-amr1} and \ref{eg-heart-amr2}  we show both of the presented examples with the corresponding AMRs. We can see the tight connection between elements of the spatial configuration and the relevant portions of the AMR and how we fit the spatial concepts as a part of the larger AMR scheme to obtain a richer representation language.

\subsection{Spatial Configuration Elements} The roles of Trajector ($tr$) and Landmarks ($lm$) are associated with  spatial entities. {\bf Trajector} roles are presented as \textit{tr=\textless id,e\textgreater} where $id$ is the identifier assigned to a $tr$ and $e$ is a spatial entity that plays this role. In our first example, \textit{the large red block} is a spatial entity with the semantic head being \textit{block}, exactly as in the AMR. A trajector can be composed \ignore{complex consisting} of multiple objects such as  
\textit{a woman and a child}. Trajectors can also be implicit in a sentence when identifiable from the context. In the sentence, \textit{Just arrived there!}, \emph{I} is implicit. In Minecraft we treat the \textit{space} itself as a trajector (Figure \ref{eg-heart-amr2}), which is again in exact alignment with the spatial AMR.

\noindent {\bf Landmarks} are represented similarly and can be complex. In the sentence
\textit{The  block is in between the sphere and the cube.},
\textit{the sphere and the cube} is a complex landmark that involves two objects.  

\noindent{\bf Spatial indicators} are represented as
\textit{ \textless id,span,{prop}\textgreater}.\ignore{These roles are defined inside the configuration.} Each indicator participates in one configuration (unless the configuration describes a path in which we case we expect multiple spatial indicators corresponding to different segments of the path) and its role and properties are described therein. The properties are represented by a name and a span but with different values such as $\{metric,degree, distance, direction, region\}$. In Configuration 2 in Table \ref{eg-heart-tab}, \textit{in\_between} indicates the relation between the \textit{spaces (e3)} and the \textit{blocks (e1,e2)}. 
and \textit{5 spaces} is the metric property for \textit{from}.

\noindent {\bf Motion indicator} is represented as \textit{ \textless id,span,{prop}\textgreater}. An example property is \textit{speed}. 
When the configuration includes a {\bf path} , particularly for the dynamic spatial descriptions or fictive motion (\textit{The river meanders through the valley.}),  the path is described using a set of landmarks and spatial indicators: 
 \textit{path= \textless\{(lm,sp,path-part)\}, prop\textgreater}. Here \textit{prop} is the property associated with the path. For example, in  Table \ref{eg-tower-tab}, \textit{Orientation} with \textit{diagonally} as its value is a property of the path. For the \textbf{path} variable, each spatial indicator which is associated with a $lm$ in the path indicates which part of the path it is referring to, where \textit{path-part} $\in\{begin,end,middle,whole\}$.
Each of these path-part values are illustrated in the example   \textit{I am going from DC (begin) to NY (end) through PHL (middle) along I-78 (whole)}. In Table \ref{eg-tower-tab}, the path includes two landmarks: \ignore{\textit{top of the blue column}}one is the beginning of the path and \ignore{\textit{top of the yellow column}}the other is the end. In Table \ref{eg-heart-tab}, the beginning of the path is implicit.
Complex configurations having a path with multiple landmarks can have multiple {\bf frames-of-reference}s (FoR). Thus, FoR is represented with respect to a landmark, as
\textit{\textless lm, value\textgreater} where \textit{value $\in$ \{intrinsic,relative,absolute\}}. \textit{Intrinsic FoR (object-centered)} activates an internal axis or part of the object of the spatial indicator, e.g., \emph{top of the building}. \emph{Relative FoR (viewpoint-centered)} uses another spatial entity to anchor the location or orientation of the current object's spatial description, e.g., \emph{my left} and \emph{your right}. \emph{Absolute FoR (geo-centered)} is a fixed FoR, e.g., \emph{North}\ignore{and \emph{South}}. 
 As an example of a configuration having multiple FoRs with different values, consider : \textit{I walked from the center of the park to the left of the statue.}
In the path for this configuration, \ignore{ this is one configuration but} the first landmark has an intrinsic FoR while the second landmark has a relative FoR. In contrast to the FoR, the {\bf viewer} is unique in a single configuration and is represented as a value $v\in\{first-person,second-person,third-person\}$~\cite{thora2011,W18-4704}.\ignore{ Unless there are no relative FoRs in a configuration, we need to know the value of the viewer to interpret the spatial meaning of the sentence, unambiguously.} A configuration describes at most one \ignore{basic/complex} trajector based on at most one basic/complex landmark or a path. It
   includes various semantic aspects that can make it specific enough for the visualization of a single spatial relation. Therefore, when there is more than one spatial indicator (except for a path which is a complex landmark with multiple spatial indicators), we have one configuration for each spatial indicator. Each configuration will have at most one motion indicator.  

\ignore{
 \begin{figure}
	\centering
	\includegraphics[width=0.5\textwidth]{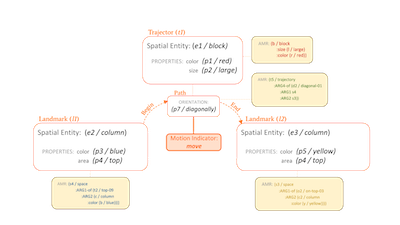}
	\caption
	{Graphical representation of the trajector, path and the motion indicator, also see Table \ref{tab:2filternum}.}
\end{figure}}
 \ignore{\subsection{Forming Configurations}}

\subsection{Qualitative Types}

 Fine-grained spatial relation semantics can be represented using specialized linguistically motivated ontologies such as General Upper Model \ignore{(GUM)}~\cite{LinOntAi}. Another approach has been mapping to formal spatial knowledge representation models that are independent from natural language~\cite{SpatialML08,LREC,iso-space2011}. 
The latter provides a more tractable formalism for spatial reasoning. The spatial formal models are divided into three categories: \textit{topological}, \textit{directional} and \textit{distal}, and for each category \ignore{or a combination of them }specific formalisms have been invented~\cite{citeulike:4171638,spcalcom}. We use these\ignore{three} qualitative types\ignore{at a high level}. Directional systems can be relative or absolute, distal systems can be qualitative or quantitative and topological models can follow various formalisms such as \ignore{region connection calculus,} RCC8, RCC5, etc~\cite{RaCC92}. Though these formalisms have the advantage of providing reasoning systems, using them is challenging due to their differences in the level of specificity compared to natural language. Thus, we leave open the option of plugging in a fine-grained formalism in our representation by representing the qualitative type as a pair of general type and formal meaning, \textit{QT=$<$G-type,F-meaning$>$}, where \textit{G-type} $\in$ \{\textit{direction,distance,topology}\} and F-meaning will be a task-specific \ignore{domain-dependent} formalism. \ignore{As in the previous SpRL scheme, }We allow the assignment of multiple \textit{G-type} and \textit{F-type} to a single configuration to cover the gap between the level of specificity of language and formal models.

\section{Automated Parsing}
\begin{figure}
	\centering
	\includegraphics[width=0.5\textwidth]{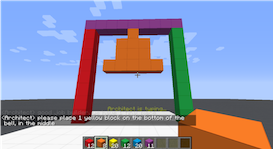}
	\caption
	{please place 1 yellow block on the bottom of the bell, in the middle.}
	\label{bell}
\end{figure}

In this section, we demonstrate that the extended AMR scheme is learn-able by training a parser on the annotated examples. We present results for the state-of-the-art AMR parser \cite{stog}(STOG parser) as a baseline for future follow-up works to parse the natural language surface form into the AMR format. These are preliminary results with a scope for further improvement in the future. The statistics for the data split are shown in Table \ref{table:sent-stats}. We achieved an F1 score (calculated through triplet matches) of 66.24$\%$ on the test set after training on the filtered training set and validating on the filtered dev set. The parser is trained from scratch instead of relying on a pre-trained version of it, since the domain of LDC2017T10 data which STOG parser was reported on is significantly different than the Minecraft data --- we found several preliminary fine-tuning results are not as good as the version trained from scratch. Here we show the predicted AMR output for the ``bell'' construction from the Minecraft dataset (Figure \ref{bell}) , which is one of the more challenging cases:
\begin{table}[t]
\centering
\small
\begin{tabular}{|l|c|c|}
\cline{2-3}
\multicolumn{1}{c|}{} & total & used \\ \hline
train & 7954 & 4850\\
\hline
dev & 933 & 604\\
\hline
test & 862 & 583\\
\hline
\end{tabular}
\caption{Total number of sentences and the number of sentences used for training, validation and test purposes \label{table:sent-stats}}
\end{table}

\begin{verbatim}
(vv1 / place-01
    :ARG0 (vv3 / you)
    :ARG1 (vv4 / block
      :ord (vv5 / ORDINAL_ENTITY_1
        :range-start (vv6 / thing
          :ARG1-of (vv7 / bottom-03
            :ARG2 (vv8 / bell))
          :ARG1-of (vv9 / middle-01
            :ARG2 (vv10 / bell))
          :ARG1-of (vv11 / middle-01
            :ARG1 vv4))))
    :mode imperative)
\end{verbatim}
which predicts correctly most of the important components of the actual gold AMR.
\begin{verbatim}
(p / place-01 :polite + :mode imperative
    :ARG0 (y / you)
    :ARG1 (b / block :quant 1 
                     :color(y2/yellow))
    :ARG2 (s / space
      :ARG1-of (m / middle-01
        :ARG2 (c / composite-entity
          :ARG1-of (b4 / bottom-03
            :ARG2 (b3 / bell))))))
\end{verbatim}

\section{Universal Applicability of \textsc{spatial configuration} scheme}
\begin{figure}[htp]
\centering
\begin{minipage}[b]{.45\textwidth}
\includegraphics[width=\textwidth]{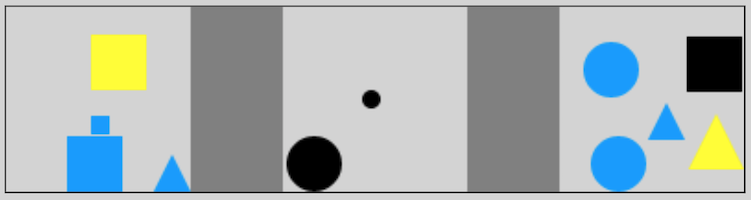}
\caption{NLVR example  : There is a blue square closely touching the bottom of a box.}\label{nlv}
\end{minipage}\qquad

\begin{minipage}[b]{.45\textwidth}
\includegraphics[width=\textwidth]{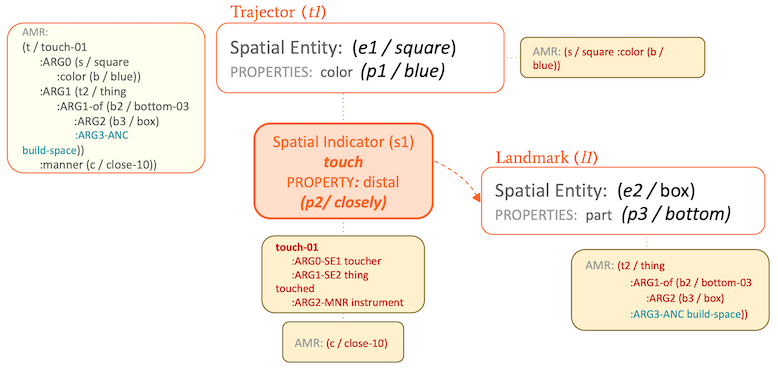}
\caption{Graphical Representation of the components of the spatial configuration with the aligned AMR for Figure \ref{nlv} :\textit{There is a blue square closely touching the bottom of a box.} (zoom in for more clarity)}\label{nl-amr}
\end{minipage}
\end{figure}

We emphasize that although we have annotated the Minecraft corpus with spatial AMR to highlight the efficacy in one extremely spatially-challenging domain, we can use the proposed scheme to represent the spatial aspects of any domain. For illustration, we present an example from the NVLR dataset (Figure \ref{nlv}), which is a TRUE statement: There is a blue square closely touching the bottom of a box. We also show the detailed annotation in Table \ref{nl} and the graphical representation with an aligned AMR in Figure \ref{nl-amr}.

\begin{figure}
\centering
\begin{minipage}[b]{.2\textwidth}
\includegraphics[width=\textwidth]{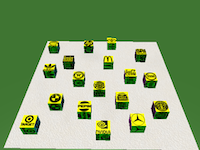}

\end{minipage}\qquad
\hfill
\begin{minipage}[b]{.2\textwidth}
\includegraphics[width=\textwidth]{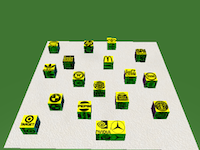}

\end{minipage}
\caption{Before (top) and After (bottom) screenshots of the execution of the instruction : \textit{Move the Mercedes block to the right of the Nvidia block.} }\label{blox}
\par
\begin{minipage}[b]{.45\textwidth}
\includegraphics[width=\textwidth]{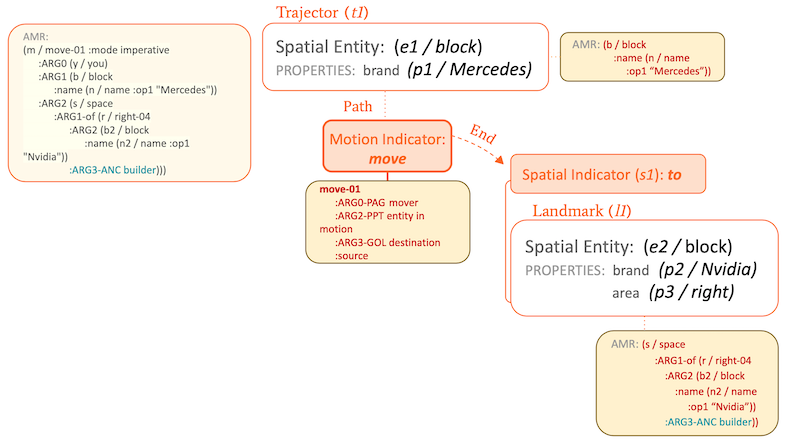}
\caption{Graphical Representation of the components of the spatial configuration with the aligned AMR for Figure \ref{blox} : \textit{Move the Mercedes block to the right of the Nvidia block.}  (zoom in for more clarity) }\label{bl-amr}
\end{minipage}

\end{figure}

 We also present another example from BLOCKS (Figure \ref{blox}): Move the Mercedes block to the right of the Nvidia block. The annotation is shown in Table \ref{bl} and the graphical representation with aligned AMR in Figure \ref{bl-amr}.

\newcolumntype{R}{>{\RaggedRight\arraybackslash}X}

\begin{table}[htbp]
    
    \centering
    \small
    %\begin{tabular}{ll}
    \begin{tabular}{|c|c|p{3.5cm}|}
                    
                     \hline
                    \multicolumn{3}{|c|}{Spatial Entities} \\ \hline
                    id & head & \makecell{properties \\\{ \textit{name = span} \}}  \\ \hline
                    e1 & square & \makecell{\{ col=blue \}} \\ \hline
                    e2& box & \makecell{\{ part-of=bottom \} } \\ \hline
                   
                \end{tabular}
         \newline

        \begin{tabular}{|c|p{3.2cm}|} 
                    \hline
                     & \makecell{Configuration 1} \\ \hline
                    tr&\makecell{$<t1, e1>$} \\ \hline
                    lm& \makecell{$<l1,e2>$}\\ \hline
                    %Lm&$<l2, e3>$\\ \hline
                    sp& \makecell{ \textless \textit{s1,touching} \\ $degree=closely$ \textgreater }  \\ \hline
                    %Sp&s2=to, []\\ \hline
                    m& \makecell{NULL}\\ \hline
                    path& \makecell{NULL}   \\ \hline   
                    FoR & \makecell{$<l1,intrinsic>$}\\ \hline
                    v& \makecell{$first-person$} \\ \hline  
                    QT& \makecell{\textit{\textless topological, EC\textgreater}} 
                    \\ \hline
                \end{tabular}    
    
   % \end{tabular}
    \caption{Relational representation of spatial entities and configurations in the sentence: \emph{There is a blue square closely touching the bottom of a box.}}
    \label{nl}
    
\end{table}

\newcolumntype{R}{>{\RaggedRight\arraybackslash}X}

\begin{table}[htbp]
    
    \centering
    \small
    %\begin{tabular}{ll}
    \begin{tabular}{|c|c|p{3.5cm}|}
                    
                     \hline
                    \multicolumn{3}{|c|}{Spatial Entities} \\ \hline
                    id & head & \makecell{properties \\\{ \textit{name = span} \}}  \\ \hline
                    e1 & block & \makecell{\{ brand=Mercedes \}} \\ \hline
                    e2& block & \makecell{\{ brand=Nvidia, \\ $area=right$ \} } \\ \hline
                   
                \end{tabular}
         \newline

       \begin{tabular}{|c|p{3.2cm}|} 
                    \hline
                     & \makecell{Configuration 1} \\ \hline
                    tr&\makecell{$<t1, e1>$} \\ \hline
                    lm&\makecell{$<l1,e2>$}\\ \hline
                    %Lm&$<l2, e3>$\\ \hline
                    sp& \makecell{\textless \textit{s1,to} \textgreater}  \\ \hline
                    %Sp&s2=to, []\\ \hline
                    m& \makecell{\textless \textit{m1,move}  \textgreater} \\ \hline
                    path& \makecell{$<implicit,begin>$ \\ $<l1,s1,end>$}\\ \hline   
                    FoR & \makecell{$<l1,relative>$}\\ \hline
                    v& \makecell{first-person} \\ \hline  
                    QT& \textit{\textless directional, relative\textgreater}
                    \\ \hline
                \end{tabular}
    
   % \end{tabular}
    \caption{Relational representation of spatial entities and configurations in the sentence: \emph{ Move the Mercedes block to the right of the Nvidia block.}}
    \label{bl}
    
\end{table}

\section{Related Representations}

The presented scheme is related to previous spatial annotation schemes such as ISO-space~\cite{iso-space2011,500427} and \ignore{spatial role labeling} SpRL~\cite{LREC,kordjamshidi-bethard-moens:2012:STARSEM-SEMEVAL,kolomiyets-EtAl:2013:SemEval-2013,bookCh2}.\ignore{These schemes expand on an earlier scheme SpatialML~\cite{SpatialML08}.}
%which has focused more on geographical and place information. 
SpRL is based on holistic spatial semantics~\cite{Holistic} and takes both cognitive-linguistic elements and spatial calculi models into account to bridge natural language and formal spatial representations that facilitate reasoning. The result is a light-weight spatial ontology compared to more extensive, linguistically motivated spatial ontologies such as GUM~\cite{LinOntAi,HoisKutz2008-fois} or type theory motivated ~\cite{dobnik2017interfacing} ones.
ISO-Space scheme considers very fine-grained properties of spatial objects based on external information and ontologies \ignore{, that can include geo-locations such as rivers, mountains or cities and countries.} Dynamic spatial relations are extensively investigated in ISO-space and its recent extensions~\cite{W18-4704}. Our proposed\ignore{semantic} representation extends ISO-space and SpRL by \ignore{introducing \textit{spatial configurations} which }decomposing complex \ignore{and nested} spatial descriptions into \ignore{a number of} modular\ignore{independent} configurations that are easier to formalize, ground and visualize. This facilitates incorporation of additional ISO-space concepts, extending our coverage\ignore{range} to more complex and varied motion events. These configurations can be composed\ignore{for building complex scenes} during grounding via shared indicators and arguments. Composition rules are domain dependent and formalizing them \ignore{in the context of}for a specific domain is future work. We represent spatial entities\ignore{and their properties} independently from the configurations. Entities can play various roles inside each configuration and the indicators can have properties. Our representation of motion along a path is more expressive than ISO-space by allowing multiple frames-of-references with respect to each constituent landmark connection in a path.\ignore{dividing the path into parts connected to various landmarks and allows multiple frames-of-references with respect to each individual landmark connection.} The spatial configuration integrates these components and we tie them to AMR, 
\ignore{thereby expanding it to be able to represent complex spatial phenomena. 
By integrating our spatial representation scheme with AMR, we are able to represent the} 
enabling representation of 
spatial semantics as a subgraph of the complete semantic representation of the sentence;
the first integration of
\ignore{. We are the first to integrate} 
spatial semantics with a rich general meaning representation scheme (AMR).
%We are pursuing a large-scale data annotation effort with the extended Propbank frames and AMR annotation to train a system for automatic annotation.

%MP Lexically based resources such as Propbank~\cite{palmer2005proposition,gildea2002necessity}, FrameNet~\cite{doi:10.1093/ijl/16.3.231}, the TRIPS parser ~\cite{ferguson1998trips},  and Abstract Meaning Representation (AMR)~\cite{banarescu2013abstract} focus on surface realizations of spatial lexical items, so one of our goals for future work will be mapping from these representations to our more canonical configuration.
%MP
\ignore{
Formalization of spatial semantics has been investigated as a secondary aspect in linguistic resources such as %Propbank~\cite{palmer2005proposition,gildea2002necessity}, 
FrameNet~\cite{doi:10.1093/ijl/16.3.231}. 

%and in more extensive abstract meaning representation (AMR) framework~\cite{banarescu2013abstract}. 

In FrameNet a generic spatial semantic frame that is independent of the language has not been the center of attention. Therefore, spatial concepts are scattered across the semantic frames. In contrast,
\ignore{to this view, spatial annotation schemes} 
we consider spatial semantics as a pivot and elaborate the different ways it is expressed in various lexical expressions and linguistic constructs.
Generic natural language semantic parsers such as the TRIPS parser~\cite{ferguson1998trips} have also considered conceptualization over spatial semantics and parsing spatial constructs using (mostly) context-free grammars and their underlying ontologies. 
However, forming spatial configurations with context-free parsers is very hard if not impossible because the concepts of each configuration are context-sensitive and do not always follow a parsable structure. Some of the concepts are beyond the linguistic constructs and are related to world knowledge of the spatial roles, as in whether an object has an intrinsic top.
\ignore{is the knowledge beyond the utterance itself though it is essential in interpreting the spatial meaning.}  Similarly, the qualitative type of the configuration depends on the types of objects, as in \ignore{involved in the relationship such as} 
\emph{crack in vase} versus \emph{flower in vase}. Such semantic gaps can be disambiguated and learned from data. 

Mapping the language to spatial configurations with qualitative types is aimed at supporting reasoning based on the types, independently of the actual objects and lexical forms, to improve the generalizability of the learning models.}

\section{Conclusion}
We presented a new spatial representation language and showed how it can compactly represent the spatial aspect of complex sentences that were challenging for existing schemes. We also integrated the scheme with AMR to achieve a richer meaning representation language. Further, we annotated sentences from the Minecraft corpus and trained an automatic parser to convert sentences into the extended AMR. The spatial configuration schema gives a succinct way to represent the spatial aspects of the full AMR annotation which we depict through examples. %We hope
The adoption of the spatial representation scheme and the extended AMR %will be adopted 
as general-purpose tools by the research community can be beneficial particularly in spatially involved domains.

\ignore{\section{Conclusion}
\ignore{In this paper, }We propose a representation language for spatial information that captures notions useful for spatial reasoning and have extended AMR by integrating these components with it. The language design is driven by our preliminary investigations into human-machine collaboration in planning, navigation and construction. 
%Our formalism is informed by schemes presented in prior work, such as \cite{LinOntAi,SpatialML08,LREC,bookCh2,iso-space2011}. Our approach handles both static and dynamic spatial configurations through a simple representation language that is designed to support composition of spatial configurations. It can also deal with components of spatial configurations even when they are complex, e.g., complex spatial entities appearing as a trajector or a landmark and are described by spatial properties. 
\ignore{We leave the domain-dependent task of formalizing composition of configurations \ignore{based on the  shared entities or domain specific rules between spatial indicators, }as future work.} We are pursuing a large-scale data annotation effort with the extended Propbank frames and AMR annotation to train a system for automatic annotation.
%\ab{R3 alluded to some open questions regarding complications in compositionality of configurations in the 2nd weakness, we can mention them in the future work.}

}

\vspace{0.5cm}
\textbf{Acknowledgements}: This work was supported by Contract W911NF-15-1-0461 with the US Defense Advanced Research Projects Agency (DARPA) and the Army Research Office (ARO) and by NSF CAREER award \#1845771. Approved for Public Release, Distribution Unlimited. The views expressed are those of the authors and do not reflect the official policy or position of the Department of Defense or the U.S. Government.

\twocolumn

\section{References}
\bibliographystyle{lrec}
\bibliography{lrec2020W-xample-kc}

\begin{thebibliography}{}

\bibitem[\protect\citename{Banarescu \bgroup et al.\egroup
  }2013]{banarescu2013abstract}
Banarescu, L., Bonial, C., Cai, S., Georgescu, M., Griffitt, K., Hermjakob, U.,
  Knight, K., Koehn, P., Palmer, M., and Schneider, N.
\newblock (2013).
\newblock Abstract meaning representation for sembanking.
\newblock In {\em Proceedings of the 7th Linguistic Annotation Workshop and
  Interoperability with Discourse}, pages 178--186.

\bibitem[\protect\citename{Bateman \bgroup et al.\egroup }2010]{LinOntAi}
Bateman, J.~A., Hois, J., Ross, R., and Tenbrink, T.
\newblock (2010).
\newblock A linguistic ontology of space for natural language processing.
\newblock {\em Artificial Intelligence}, 174(14):1027--1071.

\bibitem[\protect\citename{Bisk \bgroup et al.\egroup }2018]{bisk2018learning}
Bisk, Y., Shih, K.~J., Choi, Y., and Marcu, D.
\newblock (2018).
\newblock Learning interpretable spatial operations in a rich 3d blocks world.
\newblock In {\em Thirty-Second AAAI Conference on Artificial Intelligence}.

\bibitem[\protect\citename{Bonn \bgroup et al.\egroup }2019]{bonn2019spatial}
Bonn, J., Palmer, M., Cai, J., and Wright-Bettner, K.
\newblock (2019).
\newblock Spatial amr: Expanded spatial annotation in the context of a grounded
  minecraft corpus.
\newblock manuscript submitted for publication.

\bibitem[\protect\citename{Dobnik and Cooper}2017]{dobnik2017interfacing}
Dobnik, S. and Cooper, R.
\newblock (2017).
\newblock Interfacing language, spatial perception and cognition in type theory
  with records.
\newblock {\em Journal of Language Modelling}, 5(2):273--301.

\bibitem[\protect\citename{Gildea and Palmer}2002]{gildea2002necessity}
Gildea, D. and Palmer, M.
\newblock (2002).
\newblock The necessity of parsing for predicate argument recognition.
\newblock In {\em Proceedings of the 40th Annual Meeting on Association for
  Computational Linguistics}, pages 239--246. Association for Computational
  Linguistics.

\bibitem[\protect\citename{Goldman \bgroup et al.\egroup }2018]{P18-1168}
Goldman, O., Latcinnik, V., Nave, E., Globerson, A., and Berant, J.
\newblock (2018).
\newblock Weakly supervised semantic parsing with abstract examples.
\newblock In {\em Proceedings of the 56th Annual Meeting of the Association for
  Computational Linguistics (Volume 1: Long Papers)}, pages 1809--1819.
  Association for Computational Linguistics.

\bibitem[\protect\citename{Hois and Kutz}2008]{HoisKutz2008-fois}
Hois, J. and Kutz, O.
\newblock (2008).
\newblock {C}ounterparts in {L}anguage and {S}pace - {S}imilarity and
  \mbox{$\mathcal{S}$-Con}\-nec\-tion.
\newblock In Carola Eschenbach et~al., editors, {\em {P}roceedings of the
  {I}nternational {C}onference on {F}ormal {O}ntology in {I}nformation
  {S}ystems {(FOIS-2008)}}, pages 266--279, Amsterdam. {IOS} {P}ress.

\bibitem[\protect\citename{Hu \bgroup et al.\egroup }2017]{Hu2017LearningTR}
Hu, R., Andreas, J., Rohrbach, M., Darrell, T., and Saenko, K.
\newblock (2017).
\newblock Learning to reason: End-to-end module networks for visual question
  answering.
\newblock {\em 2017 IEEE International Conference on Computer Vision (ICCV)},
  pages 804--813.

\bibitem[\protect\citename{Hudson and Manning}2019]{hudson2019gqa}
Hudson, D.~A. and Manning, C.~D.
\newblock (2019).
\newblock Gqa: a new dataset for compositional question answering over
  real-world images.
\newblock {\em arXiv preprint arXiv:1902.09506}.

\bibitem[\protect\citename{Kolomiyets \bgroup et al.\egroup
  }2013]{kolomiyets-EtAl:2013:SemEval-2013}
Kolomiyets, O., Kordjamshidi, P., Moens, M.-F., and Bethard, S.
\newblock (2013).
\newblock Semeval-2013 task 3: Spatial role labeling.
\newblock In {\em Second Joint Conference on Lexical and Computational
  Semantics (*SEM), Volume 2: Proceedings of the Seventh International Workshop
  on Semantic Evaluation (SemEval 2013)}, pages 255--262, Atlanta, Georgia,
  USA, June.

\bibitem[\protect\citename{Kordjamshidi \bgroup et al.\egroup }2010]{LREC}
Kordjamshidi, P., {van Otterlo}, M., and Moens, M.-F.
\newblock (2010).
\newblock Spatial role labeling: task definition and annotation scheme.
\newblock In Nicoletta Calzolari, et~al., editors, {\em Proceedings of the
  Seventh Conference on International Language Resources and Evaluation
  ({LREC}'10)}, pages 413--420.

\bibitem[\protect\citename{Kordjamshidi \bgroup et al.\egroup
  }2012]{kordjamshidi-bethard-moens:2012:STARSEM-SEMEVAL}
Kordjamshidi, P., Bethard, S., and Moens, M.-F.
\newblock (2012).
\newblock {SemEval}-2012 task 3: Spatial role labeling.
\newblock In {\em Proceedings of the First Joint Conference on Lexical and
  Computational Semantics: Proceedings of the Sixth International Workshop on
  Semantic Evaluation {(SemEval)}}, volume~2, pages 365--373.

\bibitem[\protect\citename{Kordjamshidi \bgroup et al.\egroup }2017]{bookCh2}
Kordjamshidi, P., van Otterlo, M., and Moens, M.-F.
\newblock (2017).
\newblock Spatial role labeling annotation scheme.
\newblock In N.~Ide James~Pustejovsky, editor, {\em Handbook of Linguistic
  Annotation}. Springer Verlag.

\bibitem[\protect\citename{Krishnaswamy \bgroup et al.\egroup }2019]{JamesSpQL}
Krishnaswamy, N., Friedman, S., and Pustejovsky, J.
\newblock (2019).
\newblock Combining deep learning and qualitative spatial reasoning to learn
  complex structures from sparse examples with noise.
\newblock In {\em Proceedings of the thirty-third {AAAI} Conference on
  Artificial Intelligence}.

\bibitem[\protect\citename{Lee \bgroup et al.\egroup }2018]{W18-4704}
Lee, K., Pustejovsky, J., and Bunt, H.
\newblock (2018).
\newblock The revision of iso-space,focused on the movement link.
\newblock In {\em Proceedings 14th Joint ACL - ISO Workshop on Interoperable
  Semantic Annotation}, pages 35--44. Association for Computational
  Linguistics.

\bibitem[\protect\citename{Liu \bgroup et al.\egroup }2009]{spcalcom}
Liu, W., Li, S., and Renz, J.
\newblock (2009).
\newblock Combining {RCC}-8 with qualitative direction calculi: algorithms and
  complexity.
\newblock In {\em IJCAI}.

\bibitem[\protect\citename{Mani \bgroup et al.\egroup }2008]{SpatialML08}
Mani, I., Hitzeman, J., Richer, J., Harris, D., Quimby, R., and Wellner, B.
\newblock (2008).
\newblock {Spatial{M}{L}}: annotation scheme, corpora, and tools.
\newblock In N.~Calzolari, et~al., editors, {\em Proceedings of the Sixth
  International Language Resources and Evaluation ({LREC})}. European Language
  Resources Association (ELRA).

\bibitem[\protect\citename{Narayan-Chen \bgroup et al.\egroup
  }2019]{narayan-chen-etal-2019-collaborative}
Narayan-Chen, A., Jayannavar, P., and Hockenmaier, J.
\newblock (2019).
\newblock Collaborative dialogue in {M}inecraft.
\newblock In {\em Proceedings of the 57th Annual Meeting of the Association for
  Computational Linguistics}, pages 5405--5415, Florence, Italy, July.
  Association for Computational Linguistics.

\bibitem[\protect\citename{Palmer \bgroup et al.\egroup
  }2005]{palmer2005proposition}
Palmer, M., Gildea, D., and Kingsbury, P.
\newblock (2005).
\newblock The proposition bank: An annotated corpus of semantic roles.
\newblock {\em Computational linguistics}, 31(1):71--106.

\bibitem[\protect\citename{Pustejovsky \bgroup et al.\egroup
  }2011]{iso-space2011}
Pustejovsky, J., Moszkowicz, J., and Verhagen, M.
\newblock (2011).
\newblock {ISO-Space}: The annotation of spatial information in language.
\newblock In {\em ACL-ISO International Workshop on Semantic Annotation
  (ISA'6)}.

\bibitem[\protect\citename{Pustejovsky \bgroup et al.\egroup }2015]{500427}
Pustejovsky, J., Kordjamshidi, P., Moens, M.-F., Levine, A., Dworman, S., and
  Yocum, Z.
\newblock (2015).
\newblock Sem{E}val-2015 task 8: {S}pace{E}val.
\newblock In {\em Proceedings of the 9th International Workshop on Semantic
  Evaluation (SemEval 2015), 9th international workshop on semantic evaluation
  (SemEval 2015), Denver, Colorado, 4-5 June 2015}, pages 884--894. ACL.

\bibitem[\protect\citename{Randell \bgroup et al.\egroup }1992]{RaCC92}
Randell, D.~A., Cui, Z., and Cohn, A.~G.
\newblock (1992).
\newblock A spatial logic based on regions and connection.
\newblock In {\em Proceedings of the 3rd International Conference on the
  Principles of Knowledge Representation and Reasoning, {KR}'92}, pages
  165--176.

\bibitem[\protect\citename{Suhr \bgroup et al.\egroup }2017a]{suhr2017corpus}
Suhr, A., Lewis, M., Yeh, J., and Artzi, Y.
\newblock (2017a).
\newblock A corpus of natural language for visual reasoning.
\newblock In {\em Proceedings of the 55th Annual Meeting of the Association for
  Computational Linguistics (Volume 2: Short Papers)}, volume~2, pages
  217--223.

\bibitem[\protect\citename{Suhr \bgroup et al.\egroup }2017b]{Suhr2017ACO}
Suhr, A., Lewis, M., Yeh, J., and Artzi, Y.
\newblock (2017b).
\newblock A corpus of natural language for visual reasoning.
\newblock In {\em ACL}.

\bibitem[\protect\citename{Tenbrink and Kuhn}2011]{thora2011}
Tenbrink, T. and Kuhn, W.
\newblock (2011).
\newblock A model of spatial reference frames in language.
\newblock In Max Egenhofer, et~al., editors, {\em Conference on Spatial
  Information Theory (COSIT'11)}, pages 371--390. Springer.

\bibitem[\protect\citename{Wallgr\"{u}n \bgroup et al.\egroup
  }2007]{citeulike:4171638}
Wallgr\"{u}n, J.~O., Frommberger, L., Wolter, D., Dylla, F., and Freksa, C.
\newblock (2007).
\newblock Qualitative spatial representation and reasoning in the
  {SparQ-Toolbox}.
\newblock In Thomas Barkowsky, et~al., editors, {\em Spatial Cognition V
  Reasoning, Action, Interaction}, volume 4387, chapter~3, pages 39--58.
  Springer Berlin/Heidelberg.

\bibitem[\protect\citename{Zhang \bgroup et al.\egroup }2019]{stog}
Zhang, S., Ma, X., Duh, K., and Van~Durme, B.
\newblock (2019).
\newblock {AMR Parsing as Sequence-to-Graph Transduction}.
\newblock In {\em Proceedings of the 57th Annual Meeting of the Association for
  Computational Linguistics (Volume 1: Long Papers)}, Florence, Italy, July.
  Association for Computational Linguistics.

\bibitem[\protect\citename{Zlatev}2003]{Holistic}
Zlatev, J.
\newblock (2003).
\newblock Holistic spatial semantics of {T}hai.
\newblock {\em Cognitive Linguistics and Non-Indo-European Languages}, pages
  305--336.

\end{thebibliography}

\end{document}